%% file: arxiv.tex
\documentclass{article}
\usepackage{times}
\usepackage{natbib}
\usepackage{amsmath,amsfonts,amsthm,amssymb}
\usepackage{hyperref,url}
\usepackage{import,subfiles}
\usepackage{tabularx}
\usepackage{algorithm}
\usepackage{algorithmic}
\usepackage{multirow}
\usepackage{fullpage}
\usepackage{paralist}
\usepackage{enumitem}
\usepackage{subfigure}
\usepackage{latex-defs}
\usepackage{color}
\usepackage{multirow}
\usepackage{arxiv} 
\usepackage{cancel}
\usetikzlibrary{backgrounds}
\usetikzlibrary{bayesnet}
\usetikzlibrary{matrix}
\usetikzlibrary{calc}

\newcommand{\apx}{the appendix}

\newcommand{\method}{{\sc RCMS} }
\newcommand{\methodp}{{\sc RCMS}}

\newcommand\theTitle{Reified Context Models}

\icmltitlerunning{\theTitle}

\begin{document}
\addtolength{\parskip}{-0.0pt}
\newcommand{\paraii}[1]{\paragraph{#1}}

\twocolumn[
  \icmltitle{\theTitle}

\icmlauthor{Jacob Steinhardt}{jsteinhardt@cs.stanford.edu}
\icmlauthor{Percy Liang}{pliang@cs.stanford.edu}
\icmladdress{Stanford University,
             353 Serra Street, Stanford, CA 94305 USA}

\vskip 0.3in
]

\begin{abstract}
A classic tension exists between exact inference in a simple model and
approximate inference in a complex model.
The latter offers expressivity and
thus accuracy, but the former provides coverage of the space,
an important property for 
confidence estimation and learning with indirect supervision.
In this work, we introduce a new approach, \emph{reified context models},
to reconcile this tension.
Specifically, we let the amount of context (the arity of the factors in a graphical model) 
be chosen ``at run-time'' by reifying it---that is,
letting this choice itself be a random variable inside the model.
Empirically, we show that our approach obtains expressivity and coverage
on three natural language tasks.

\end{abstract}

\subfile{introduction.stex}

\subfile{example.stex}

\subfile{tasks.stex}

\subfile{framework.stex}

\subfile{precision.stex}

\subfile{accuracy.stex}

\subfile{context.stex}

\subfile{discussion.stex}

\bibliography{icml2015}
\bibliographystyle{icml2015}
\clearpage
\appendix

\newcommand{\ED}{\small{\texttt{E}}}
\newcommand{\MD}{\small{\texttt{M}}}
\newcommand{\forward}{\mathrm{forward}}
\newcommand{\act}{\mathrm{active}}
\newcommand{\lcs}{\operatorname{lcs}}
\newcommand{\len}{\operatorname{len}}
\section{Implementation Details}
{Recall that to implement the \method method, we need to perform 
the following steps:
\begin{enumerate}
\item Let $\tilde{\sC}_i = \{c_{i-1} \times \{y_i\} \mid c_{i-1} \in \sC_{i-1}, y_i \in \sY_i\}$.
\item Compute what the mass of each element of $\tilde{\sC}_i$ would be if we used $q_{\theta}^i$ 
      as the model and $\tilde{\sC}_i$ as the collection of contexts.
\item Let $\sC_i$ be the $B$ elements of $\tilde{\sC}_i$ with highest mass, 
      together with the set $\sY_{1:i}$.
\end{enumerate}
As in Section~\ref{sec:adaptive}, each context in $\sC_i$ can be represented 
by a string $s_{1:i}$, where $s_j \in \sY_j \cup \{\star\}$. We 
will also assume an arbitrary ordering on $\sY_j \cup \{\star\}$ that 
has $\star$ as its maximum element.

In addition, we use two datatypes: \ED\ (for ``expand''), which keeps track of elements 
of $\tilde{\sC}_i$, and \MD\ (for ``merge''), which keeps track of elements 
of $\sC_i$. More precisely, if $c_{i-1}$ is represented by an object $m_{i-1}$ of 
type \MD, then $\ED(m_{i-1},y_i)$ represents $c_{i-1} \times \{y_i\}$; and 
$\MD(\ED(m_{i-1},y_i))$ represents $c_{i-1} \times \{y_i\}$ as well, with the 
distinction that it is a member of $\sC_i$ rather than $\tilde{\sC_i}$. The distinction 
is important because we will also want to merge smaller contexts into objects of 
type \MD. For both \ED\ and \MD\ objects, we maintain a field $\len$, which is the 
length of the suffix of $y_{1:i}$ that is specified (e.g., if an object represents 
$\sY_{1:3} \times \{y_{4:5}\}$, then its $\len$ is $2$).

Throughout our algorithm, we will maintain $2$ invariants:
\begin{itemize}
\item $\tilde{\sC}_i$ and $\sC_i$ will be sorted lexicographically (e.g. based first 
      on $s_i$, then $s_{i-1}$, etc.)
\item A list $\widetilde{\lcs}_i$ of length $\len(\tilde{\sC}_i)$ is maintained, such 
      that the longest common suffix of $\tilde{\sC}_i[a]$ and $\tilde{\sC}_i[b]$ 
      is $\min_{c \in [a,b)} \widetilde{\lcs}_i[c]$. A similar list $\lcs_i$ is 
      maintained for $\sC_i$.
\end{itemize}

\paragraph{Step 1.} To perform step $1$ above, we just do:
\begin{algorithmic}
\STATE $\tilde{\sC}_i = []$
\FOR{$j = 0$ {\bfseries to} $\len(\sY_i)-1$}
  \FOR{$k = 0$ {\bfseries to} $\len(\sC_{i-1})-1$}
    \IF{$k+1 < \len(\sC_{i-1})$}
    \STATE $\widetilde{\lcs}_i$.append($\lcs_{i-1}[k]+1$)
    \ELSE
    \STATE $\widetilde{\lcs}_i$.append($0$)
    \ENDIF
    \STATE $\tilde{\sC}_i$.append(\ED($\sC_i[k]$, $\sY_i[j]$))
  \ENDFOR
\ENDFOR
\end{algorithmic}
The important observation is that if two sequences end in the same 
character, their $\lcs$ is one greater than the $\lcs$ of the remaining 
sequence without that character; and if they end in different characters, 
their $\lcs$ is $0$.

Each \ED\ keeps track of a forward score, defined as 
\[ \ED(m, y).\forward = m.\forward \times \exp(\theta^{\top}\phi(m,y)). \]

\paragraph{Step 2.} For step $2$, we find the $B$ elements $\tilde{c}$ 
of $\tilde{\sC}_i$ with the largest forward score; we set a flag 
$\tilde{c}.\act$ to true for each such $\tilde{c}$.

\paragraph{Step 3.} 
\newcommand{\stack}{\operatorname{stack}}
Step $3$ contains the main algorithm challenge,
which is to efficiently 
merge each element of $\tilde{\sC}_i$ into its least ancestor in $\sC_i$. If 
we think of $\sC_i$ as a tree (as in Figure~\ref{fig:context-illustration}), 
we can do this by essentially performing a depth-first-search of the tree. The 
DFS goes backwards in the lexicographic ordering, so we need to reverse the 
lists $\sC_i$ and $\lcs_i$ at the end.
\begin{algorithmic}
\STATE \Comment{merge and update $\lcs$}
\STATE $\stack = []$
\STATE $\sC_i = []$
\STATE $\lcs_i = []$
\STATE $l \gets \infty$
\FOR{$j = \len(\tilde{\sC}_i)-1$ {\bfseries to} $0$}
  \STATE $l \gets \min(l, \widetilde{\lcs}_i[j])$
  \WHILE{$l < \stack[-1].\len$}
    \STATE \Comment{then current top of stack is not an ancestor of $\tilde{\sC}_i[j]$}
    \STATE $\stack$.pop()
  \ENDWHILE
  \IF{$\tilde{\sC}_i[j].\act$}
    \STATE $m = \MD(\tilde{\sC}_i[j])$
    \STATE $\lcs_i$.append($l$)
    \STATE $\sC_i$.append($m$)
    \STATE $\stack$.push($m$)
    \STATE $l \gets \infty$
  \ELSE
    \STATE \Comment{merge $\tilde{\sC}_i[j]$ into its least ancestor}
    \STATE $\stack[-1].\operatorname{absorb}(\tilde{\sC}_i[j])$
  \ENDIF
\ENDFOR
\STATE $\lcs_i$.reverse()
\STATE $\sC_i$.reverse()
\end{algorithmic}
If $m \in \sC_i$ has absorbed elements $e_1,\ldots,e_k$, then 
we compute $m.\forward$ as $\sum_{j=1}^k e_j.\forward$.

\newcommand{\back}{\operatorname{backward}}
After we have constructed $\sC_1,\ldots,\sC_{i-1}$, we also need 
to send backward messages for inference. If $e \in \tilde{\sC}_i$ 
is merged into $m \in \sC_i$, then $e.\back = m.\back$. If 
$m \in \sC_i$ expands to $\ED(m,y)$ for $y \in \sY_{i+1}$, then 
$m.\back = \sum_{y \in \sY_{i+1}} \ED(m,y).\back \times \exp(\theta^{\top}\phi(m,y))$. The (un-normalized) probability mass of an object is then simply 
the product of its forward and backward scores; we can compute the normalization 
constant by summing over $\sC_i$.

In summary, our method can be coded in three steps; first, during the 
forward pass of inference, we:
\begin{enumerate}
\item Expand to $\tilde{\sC}_i$ and construct $\widetilde{\lcs}_i$.
\item Sort by forward score and mark active nodes in $\tilde{\sC}_i$ for 
      inclusion in $\sC_i$.
\item Merge each node in $\tilde{\sC}_i$ into its least ancestor in $\sC_i$, 
      using a depth-first-search.
\end{enumerate}
Finally, once all of the $\sC_i$ are constructed, we perform the backward pass:
\begin{enumerate}
\item[4.] Propagate backward messages and compute the normalization constant.
\end{enumerate}
}
\section{Further Details of Experimental Setup}
\label{sec:experiment-details}
We include here a few experimental details that did not fit into the main text. 
When training with AdaGrad, we performed several stochastic gradient updates in 
parallel, similar to the approach described in \citet{recht2011hogwild} (although we 
parallelized even more aggressively at the expense of theoretical guarantees).
We also used a randomized truncation scheme to round most small coordinates 
of the gradients to zero, which substantially reduces memory usage as well as 
concurrency overhead.

For decipherment, we used absolute discounting with discount $0.25$ and smoothing 
$0.01$, and Laplace smoothing with parameter $0.01$. For the $1$st-order model, beam 
search performs better if we use Laplace smoothing instead of absolute discounting 
(though still worse than \methodp). In order to maintain a uniform experimental setup, 
we excluded this result from the main text.

For the hybrid selection algorithm in the speech experiments, we take the union 
of the beams at every step (as opposed to computing two sets of beams separately 
and then taking a single union at the end).

\section{Additional Files}
In the supplementary material, we also include the source code and datasets 
for the decipherment task. A \verb=README= is included to explain how to run 
these experiments.

\end{document}

%% file: introduction.stex
\section{Introduction}
\label{sec:introduction}

Many structured prediction tasks across natural language processing, computer vision, and computational biology
can be formulated as that of learning a distribution over outputs $y_{1:L} = (y_1, \dots,
y_L) \in \sY_{1:L}$ given an input $x$:
\begin{equation}
p_{\theta}(y_{1:L} \mid x) \propto \exp\p{\sum_{i=1}^L \theta^{\top}\phi_i(y_{1:i-1}, y_{i}, x)}.
\end{equation}
The thirst for expressive models
(e.g., where $y_i$ depends heavily on its context $y_{1:i-1}$)
often leads one down the route of approximate inference,
for example, to Markov chain Monte Carlo \cite{brooks2011handbook}, sequential Monte Carlo 
\cite{smc,particle-filter}, or beam search \cite{koehn2003statistical}.
While these methods in principle can operate on models with arbitrary 
amounts of context, they only touch a small portion of the output space $\sY_{1:L}$.
Without such \emph{coverage}, we miss out on two important but oft-neglected properties:

\begin{itemize}[itemsep=2pt,topsep=0pt,parsep=0pt,partopsep=0pt,leftmargin=12pt]
\item \textbf{precision}: In user-facing applications, it is important to 
      only predict on inputs where the system is confident, leaving hard 
      decisions to the user
      \citep{zhang2014face}.
      Lack of coverage means failing to consider all alternative outputs,
      which leads to overconfidence and poor estimates of uncertainty.

    \item \textbf{indirect supervision}:
    When only part of the output $y_{1:L}$ is observed,
    lack of coverage is even more problematic than it is in the fully-supervised setting.
    An approximate inference algorithm might not even consider the true $y$ (whereas
    one always has the true $y$ in a fully-supervised setting),
    which leads to invalid parameter updates \citep{yu2013max}.
\end{itemize}

Of course, lower-order models admit exact inference and ensure coverage,
but these models have unacceptably low expressive power.
Ideally, we would like a model that varies the amount of context in a judicious way,
allocating modeling power to parts of the input that demand it.
Therein lies the principal challenge: How can we adaptively choose the amount of context
for each position $i$ in a data-dependent way while maintaining tractability?

In this paper, we introduce a new approach, which we call \emph{reified context
models}.  The key idea is based on \emph{reification}, a general idea in logic and programming
languages, which refers to making something previously unaccessible (e.g., functions or
metadata of functions) a ``first-class citizen'' and therefore available (e.g., via lambda
abstraction or reflection) to formal manipulation.
In the probabilistic modeling setting,
we propose reifying the contexts as random variables in the model so that we
can reason over them.

\subfile{volcanic.stex}

Specifically, 
for each $i \in \{1,\ldots,L-1\}$, we maintain a 
collection $\sC_i$ of \emph{contexts}, each of which is a subset of $\sY_{1:i}$
representing what we remember about the past
(see 
Figure~\ref{fig:volcanic} for an example). 
We define a joint model over $(y,c)$,
suppressing $x$ for brevity:
\begin{equation}
\label{eq:family}
p_{\theta}(y_{1:L},c_{1:L-1}) \propto \exp\p{\sum_{i=1}^L \theta^{\top}\phi_i(c_{i-1}, y_{i})} \times \kappa(y, c),
\end{equation}
where $\kappa$ is a \emph{consistency} potential,
to be explained later.
The features $\phi_i$ now depend on the current context $c_{i-1}$, rather 
than the full history $y_{1:i-1}$.
The distribution over $(y,c)$ factorizes according to the graphical model below:
\begin{equation}
\label{pic:graphical-model}
\begin{tikzpicture}[baseline=(current  bounding  box.center)]
\tikzstyle{every node}=[minimum size=2.4em, inner sep=-0.5pt]
\def \s {1.3};
\def \t {0.85em};
\def \u {0.80em};
\node[latent] (x1) at (1*\s,1) {$Y_1$};
\node[latent] (x2) at (2*\s,1) {$Y_2$};
\node[latent] (x3) at (3*\s,1) {$Y_3$};
\node[latent] (x4) at (4*\s,1) {$Y_4$};
\node[latent] (x5) at (5*\s,1) {$Y_5$};
\node[latent] (i1) at (1.4*\s,2.5) {$C_1$};
\node[latent] (i2) at (2.4*\s,2.5) {$C_2$};
\node[latent] (i3) at (3.4*\s,2.5) {$C_3$};
\node[latent] (i4) at (4.4*\s,2.5) {$C_4$};
\node (i5) at (5*\s,2.5) {};

\factor[above=\u of x1, draw=black, color=black] {x11} {} {} {};
\factoredge[draw=black] {i1} {x11} {x1};
\factor[above=\u of x2, draw=black, color=black] {x22} {} {} {};
\factoredge[draw=black] {i2} {x22} {x2};
\factor[above=\u of x3, draw=black, color=black] {x33} {} {} {};
\factoredge[draw=black] {i3} {x33} {x3};
\factor[above=\u of x4, draw=black, color=black] {x44} {} {} {};
\factoredge[draw=black] {i4} {x44} {x4};
\factor[above=\u of x5, draw=black, color=black] {x55} {} {} {};
\factoredge[draw=black] {i1} {x22} {i2};
\factoredge[draw=black] {i2} {x33} {i3};
\factoredge[draw=black] {i3} {x44} {i4};
\factoredge[draw=black] {i4} {x55} {x5};
\end{tikzpicture}
\end{equation}
The factorization \eqref{pic:graphical-model} implies that 
the family in \eqref{eq:family} admits efficient exact 
inference via the forward-backward algorithm as long as each 
collection $\sC_i$ has small cardinality.

\paraii{Adaptive selection of context.}
Given limited computational resources, we only want to track 
contexts that are reasonably likely to contain the answer. 
We do this by selecting the context sets $\sC_i$ during a forward 
pass using a heuristic similar to beam search,
but unlike 
beam search, we achieve coverage because we are selecting 
\emph{contexts} rather than individual 
variable values. We detail our selection method, called \methodp, 
in Section~\ref{sec:adaptive}. 
We can think of selecting the $\sC_i$'s as selecting 
a model to perform inference in, with the guarantee 
that all such models will be tractable.
Our method is simple to implement; see \apx\ for implementation details.

The goal of this paper is to flesh out the framework described 
above, providing intuitions about its use and 
exploring its properties empirically. To this end, we start 
in Section~\ref{sec:tasks} by defining some tasks that motivate 
this work. In 
Sections~\ref{sec:framework} and \ref{sec:adaptive} we 
introduce reified context models formally, together with 
an algorithm, \methodp, for selecting contexts adaptively at run-time.
Sections~\ref{sec:precision}-\ref{sec:context} 
explore the empirical properties of the \method method. 
Finally, we discuss future research directions in Section~\ref{sec:discussion}.

%% file: volcanic.stex
\newcommand{\bl}[1]{\texttt{\color{blue} #1}}
\newcommand{\re}[1]{\texttt{\color{red} #1}}

\begin{figure}[b!]
\vspace{-0.15in}
\centering
\begin{tabular}{cccccccc}
\raisebox{-.5\height}{\includegraphics{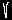}} &
\raisebox{-.5\height}{\includegraphics{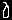}} &
\raisebox{-.5\height}{\includegraphics{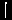}} &
\raisebox{-.5\height}{\includegraphics{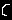}} &
\raisebox{-.5\height}{\includegraphics{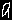}} &
\raisebox{-.5\height}{\includegraphics{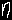}} &
\raisebox{-.5\height}{\includegraphics{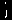}} &
\raisebox{-.5\height}{\includegraphics{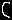}} 
\\[\normalbaselineskip]
\bl{r} & \bl{ro} & \bl{rol} & \re{*olc} 
 & \multicolumn{4}{c}{\multirow{4}{*}{
\bl{\tiny $\bullet\ \bullet\ \bullet\ \bullet\ \bullet\ \bullet\ \bullet\ \bullet\ \bullet$}
}} \\
\re{v} & \bl{ra} & \bl{ral} & \bl{***c} \\
\bl{y} & \re{*o} & \re{*ol} & \bl{***r} \\
\bl{*} & \bl{**} & \bl{***} & \bl{****} 
\end{tabular}
\caption{Illustration of our method 
for handwriting recognition. 
At each position, we keep track of a collection of 
contexts, and learn a model that factorizes with respect 
to these contexts. Each context remembers a certain amount 
of history, e.g. $\star o$ is all length two sequences
whose second character is $o$.
By using contexts at multiple levels 
of resolution, we can obtain coverage of the entire space 
while still modeling complex dependencies.}
\label{fig:volcanic}
\end{figure}

%% file: tasks.stex
\section{Description of Tasks}
\label{sec:tasks}

To better understand the motivation for our work, we present three tasks 
of interest, which are also the tasks used in our empirical evaluation later.
These tasks are word recognition (a fully supervised task), speech recognition (a weakly 
supervised task), and decipherment (an unsupervised task).
The first of these tasks is relatively easy while the latter two are harder.
We use word recognition to study the precision of our method, the other two tasks to explore 
learning under indirect supervision, and all three to understand how our algorithm selects 
contexts during training. 

\paraii{Word recognition}

The first task is the word recognition task from \citet{kassel1995comparison}; we use the 
``clean'' version of the dataset as in \citet{weiss2012structured}. This contains 
$6,876$ examples, split into $10$ folds (numbered $0$ to $9$); we used fold $1$ for testing and 
the rest for training.
Each input is a sequence of $16 \times 8$ binary images of characters; the output is the 
word that those characters spell. The first character is omitted due to capitalization issues. 
Since this task ended up being too easy as given, we downsampled each image to 
be $8 \times 4$ (by taking all pixels whose coordinates were both odd). An example input and 
output is given below:
\begin{center}
\begin{tabular}{ccccccccccc}
input $x$ &
\includegraphics{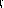} &
\includegraphics{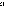} &
\includegraphics{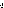} &
\includegraphics{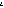} &
\includegraphics{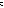} &
\includegraphics{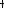} &
\includegraphics{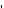} &
\includegraphics{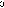} &
\includegraphics{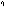} &
\includegraphics{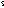} \\
output $y$ &
r &
o &
j &
e &
c &
t &
i &
o &
n &
s
\end{tabular}
\end{center}
Each individual image is too noisy to interpret
in isolation, and so leveraging 
the context of the surrounding characters is crucial 
to achieving high accuracy.

\paraii{Speech recognition}
Our second task is from the Switchboard speech transcription 
project \citep{greenberg1996insights}.
The dataset consists of $999$ utterances, split into two 
chunks of sizes $746$ and $253$; we used the latter chunk as a test set.
Each utterance is a phonetic input and textual output: 
\begin{center}
\begin{tabular}{cl}
input $x$     & h\# y ae ax s w ih r dcl d h\# \\
latent $z$    & (alignment) \\
output $y$    & yeah\_it's\_weird 
\end{tabular}
\end{center}
The alignment between the input and output is unobserved.

The average input length is $26$ phonemes, or $2.5$ seconds of speech.
We removed most punctuation from the output, except for spaces, apostrophes, dashes, and dots.

\paraii{Decipherment}

Our final task is a decipherment task similar to that described in
\citet{nuhn2014homophonics}. 
In decipherment, one is given a large amount of plain text and a smaller amount 
of cipher text; the latter is drawn from the same distribution as the former but is 
then passed through a $1$-to-$1$ substitution cipher. For instance, the plain text 
sentence ``I am what I am'' might be enciphered as ``13 5 54 13 5'':
\begin{center}
\begin{tabular}{cccccc}
latent $z$ & I & am & what & I & am \\
output $y$ & 13 & 5 & 54 & 13 & 5
\end{tabular}
\end{center}
The task is to reverse the substitution cipher, 
e.g. determine that $13 \mapsto I$, $5 \mapsto am$, etc.

We extracted a dataset from the English Gigaword corpus
\cite{graff2003} by finding the $500$ most common words 
and filtering for sentences that only contained those words. 
This left us with $24,666$ utterances, of which $2,000$ were enciphered 
and the rest were left as plain text.

Note that this task is unsupervised, but we can hope to gain information about the cipher 
by looking at various statistics of the plaintext and ciphertext. For instance, a very 
basic idea would be to match words based on their frequency. This alone doesn't work very well, 
but by considering bigram and trigram statistics we can do much better.

%% file: framework.stex
\section{Reified Context Models}
\label{sec:framework}

We now formally introduce reified context models.
Our setup is structured prediction, where
we predict an output $(y_1,\ldots,y_L) \in \sY_1 \times \cdots \times \sY_L$; 
we abbreviate these as $y_{1:L}$ and $\sY_{1:L}$.
While this setup assumes a generative model, we can easily handle 
discriminative models as well as a variable length $L$; we ignore 
these extensions for simplicity.

Our framework reifies the idea of context as a tool for efficient inference. 
Informally, a \emph{context} for $\sY_i$ is information that we remember about 
$\sY_{1:i-1}$. In our case, a context $c_{i-1}$ is a subset of $\sY_{1:i-1}$, which 
should contain $y_{1:i-1}$: in other words, $c_{i-1}$ is ``remembering'' that 
$y_{1:i-1} \in c_{i-1}$. A context set $\sC_{i-1}$ is a collection of possible values 
for $c_{i-1}$.

Formally, we define a \emph{canonical context set} $\sC_i$ to be a 
collection of subsets of $\sY_{1:i}$ satisfying two properties:\footnote{
This is similar to the definition of a \emph{hierarchical decomposition} 
from \citet{steinhardt2014filtering}. Our closure condition replaces the more 
restrictive condition that $c \cap c' \in \{c,c',\emptyset\}$.}
\begin{itemize}[itemsep=2pt,topsep=0pt,parsep=0pt,partopsep=0pt,leftmargin=12pt]
\item \textbf{coverage:} $\sY_{1:i} \in \sC_i$ 
\item \textbf{closure:} for $c, c' \in \sC_i$, $c \cap c' \in \sC_i \cup \{\emptyset\}$
\end{itemize}
An example of such a collection is given in Figure~\ref{fig:context-illustration}; as 
in Section~\ref{sec:introduction}, notation such as $\star\!\star\!a$ denotes 
the subset of $\sY_{1:3}$ where $y_3 = a$.

We refer to each element of $\sC_i$ as a ``context''. Given a sequence $y_{1:L}$, we need 
to define contexts $c_{1:L-1}$ such that $y_{1:i} \in c_i$ for all $i$. The coverage property 
ensures that \emph{some} such context always exists: we can take $c_i = \sY_{1:i}$. 

In reality, we would like to use the smallest (most precise) context $c_i$ possible; the 
closure property ensures that this is canonically defined: given a context 
$c_{i-1} \in \sC_{i-1}$ and a value $y_i \in \sY_i$, we inductively define 
$c_i = f_i(c_{i-1}, y_i)$ to be the
intersection of all $c \in \sC_i$ that contain $c_{i-1} \times \{y_i\}$, or equivalently 
the smallest such $c$.
Example evaluations of $f$ are provided
in Figure~\ref{fig:context-illustration}. Note that $y_{1:i} \in c_i$ always.

We now define a joint model over the variables $y_{1:L}$ and contexts $c_{1:L-1}$:
\begin{align*}
p_{\theta}(y_{1:L}, c_{1:L-1}) &\propto \exp\p{\sum_{i=1}^L \theta^{\top}\phi_i(c_{i-1}, y_i)} \times \kappa(y, c), 
\end{align*}
where $\kappa(y,c) \eqdef \prod_{i=2}^{L-1} \bI[c_i = f_i(c_{i-1}, y_i)]$ 
enforces consistency of contexts. 
The distribution $p_{\theta}$ factors according to \eqref{pic:graphical-model}. One 
consequence of this is that the variables $y_{1:L}$ are jointly independent given 
$c_{1:L-1}$: the contexts contain all information about interrelationships between the $y_i$.

Mathematically, the model above is similar to a hidden Markov model where $c_i$ is the 
hidden state. However, we choose the context sets \emph{adaptively}, giving us much 
greater expressive power than an HMM, since we essentially have exponentially many choices 
of hidden states (canonical context sets) to select from at runtime.

{
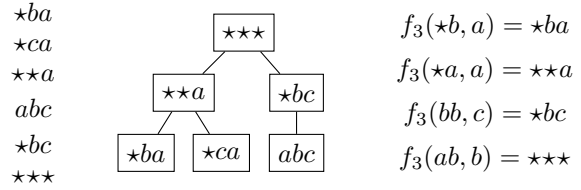
\begin{figure}
\caption{Illustration of a context set $\sC_3$. These sets form a hierarchy, 
allowing us to focus on certain specific values in $\sY_{1:3}$, while also 
allocating some resources (via the context $\star\!\star\!\star$) to model 
the rest of $\sY_{1:3}$ as well.
To the right are some example outputs of $f_3$.}
\label{fig:context-illustration}
\begin{tikzpicture}
\def \y {-0.5}
\def \x {1.0}
\matrix at (0,-0.1) [matrix of nodes,row sep=0,column 1/.style={nodes={rectangle,minimum width=3em}}]
{
$\star ba$ \\
$\star ca$ \\
$\star\!\star\!a$ \\
$abc$ \\
$\star bc$ \\
$\star\!\star\!\star$\\
};

\matrix at (6,-0.1) [matrix of nodes,row sep=0,column 1/.style={nodes={rectangle,minimum width=3em}}]
{
$f_3(\star b,a) = \star ba$ \\
$f_3(\star a,a) = \star\!\star\! a$ \\
$f_3(bb,c) = \star bc$ \\
$f_3(ab,b) = \star\!\star\!\star$ \\
};

\tikzstyle{every node}=[draw,rectangle,minimum height=1.4em]
\node[draw,rectangle] (sss) at (2.8, 0.7) {$\star\!\star\!\star$};
\node[draw,rectangle] (ssa) at (2.0, -0.1) {$\star\!\star\!a$};
\node[draw,rectangle] (sbc) at (3.5, -0.1) {$\star bc$};
\node[draw,rectangle] (sba) at (1.5, -0.9) {$\star ba$};
\node[draw,rectangle] (sca) at (2.5, -0.9) {$\star ca$};
\node[draw,rectangle] (abc) at (3.5, -0.9) {$abc$};
\draw (sss) -- (ssa);
\draw (sss) -- (sbc);
\draw (ssa) -- (sba);
\draw (ssa) -- (sca);
\draw (sbc) -- (abc);
\end{tikzpicture}
\vspace{-3.5ex}
\end{figure}
}

\paraii{Example: $2$nd-order Markov chain.}
To provide more intuition, we construct a
$2$nd-order Markov chain using our framework (we can construct
$n$th-order Markov chains in the same way).
We would like $\sC_{i}$ to ``remember'' 
the previous $2$ values, i.e. $(y_{i-1},y_{i})$.
To do this, we let $\sC_{i}$ consist of all sets of the form 
$\sY_{1:i-2} \times \{(y_{i-1},y_i)\}$; these sets fix the value 
of $(y_{i-1}, y_i)$ while allowing $y_{1:i-2}$ to vary freely.
Then $f_{i+1}(c_{i}, y_{i+1}) = \sY_{1:i-1} \times \{(y_i, y_{i+1})\}$, 
which is well-defined since $y_i$ can be determined from $c_i$.

If $|\sY_i| = V$, then $|\sC_i| = V^2$ (or $V^n$ for $n$th-order chains), 
reflecting the cost of inference in such models.

As a technical note, we also need to include $\sY_{1:i}$ in $\sC_i$ to 
satisfy the coverage condition. However, $\sY_{1:i}$ will never actually 
appear as a context, as can be seen by the preceding definition of $f$.

To finish the construction, suppose we have a family of $2$nd-order Markov chains 
parameterized as
\begin{equation}
\label{eq:hmm}
p_{\theta}(y_{1:n}) \propto \exp\p{\sum_{i=1}^L \theta^{\top}\phi_i((y_{i-2}, y_{i-1}), y_i)}.
\end{equation}
Since $\phi_i$ depends only on $(y_{i-2}, y_{i-1})$, which can be determined from 
$c_{i-1}$, we can define an equivalent function $\tilde{\phi}_i(c_{i-1}, y_i)$. 
Doing so, we recover a model family equivalent to \eqref{eq:hmm} after marginalizing 
out $c_{1:L-1}$ (since $c_{1:L-1}$ is a deterministic 
function of $y_{1:L}$, this last step is trivial).

\section{Adaptive Context Sets}
\label{sec:adaptive}

The previous section shows how to define a tractable model for any 
collection of canonical context sets $\sC_i$. We now
show how to choose such sets adaptively, at run-time. We use 
a heuristic motivated by beam search, which greedily chooses the 
highest-scoring configurations of $y_{1:i}$ based on an estimate 
of their mass. We work one level of abstraction higher, choosing 
\emph{contexts} instead of \emph{configurations}; this allows us 
to maintain coverage while still being adaptive.

Our idea has already been illustrated to an extent in Figures~\ref{fig:volcanic} and 
\ref{fig:context-illustration}: if some of our contexts are very coarse (such as 
$\star\!\star\!\star$ in Figure~\ref{fig:context-illustration}) and others are 
much finer (such as $abc$ in Figure~\ref{fig:context-illustration}), then we can 
achieve coverage of the space while still modeling complex dependencies. We will do this by allowing each context $c \in \sC_i$ 
to track a suffix of $y_{1:i}$ of arbitrary length; this contrasts with the Markov 
chain example where we always track suffixes of length $2$.

Precise contexts expose more information 
about $y_{1:i}$ and so allow more accurate modeling; however, they 
are small and $\sC_i$ is necessarily limited in size, so only a small part 
of the space can be precisely modeled in this way. We thus want
to choose contexts that focus on high probability regions.

\paraii{Our procedure.}
To do this, we define the \emph{partial model} 
\begin{equation*}
q_{\theta}^i(y_{1:i},c_{1:i-1}) \propto \exp\p{\sum_{j=1}^{i} \theta^{\top}\phi_j(c_{j-1},y_j)} \times \kappa(y,c).
\end{equation*}
We then define the context sets inductively via the following procedure, which takes as input 
a \emph{context size} $B$: 
\begin{itemize}[itemsep=2pt,topsep=0pt,parsep=0pt,partopsep=0pt,leftmargin=12pt]
\item Let $\tilde{\sC}_i = \{c_{i-1} \times \{y_i\} \mid c_{i-1} \in \sC_{i-1}, y_i \in \sY_i\}$.
\item Compute the mass of each element of $\tilde{\sC}_i$ under $q_{\theta}^i$. 
\item Let $\sC_i$ be the $B$ elements of $\tilde{\sC}_i$ with highest mass, 
      together with the set $\sY_{1:i}$.
\end{itemize}
The remaining elements of $\tilde{\sC}_i$ will effectively be merged into their least ancestor 
in $\sC_i$.
Note that each $c \in \sC_i$ fixes the value of some suffix $y_{j:i}$ of $y_{1:i}$, and 
allows $y_{1:j-1}$ to vary freely across $\sY_{1:j-1}$. Any such collection will automatically 
satisfy the closure property.

The above procedure can be performed during the forward pass of inference, and so is 
cheap computationally. Implementation details can be found in \apx. We call this procedure
\method (short for ``Reified Context Model Selection'').

\paraii{Caveat.}
There is no direct notion of an inference error in the above procedure, 
since exact inference is possible by design. An indirect notion of 
inference error is poor choice of contexts, which can lead to less accurate predictions.

\subsection{Relationship to beam search}
The idea of greedily selecting contexts based on $q_{\theta}^i$ is 
similar in spirit to beam search, an approximate inference algorithm 
that greedily selects individual values of $y_{1:i}$ based 
on $q_{\theta}^i$. More formally, beam search maintains a 
\emph{beam} $\sB_i \subseteq \sY_{1:i}$ of size $B$, constructed 
as follows:
\begin{itemize}[itemsep=2pt,topsep=0pt,parsep=0pt,partopsep=0pt,leftmargin=12pt]
\item Let $\tilde{\sB}_i = \sB_{i-1} \times \sY_i$.
\item Compute the mass of each element of $\tilde{\sB}_i$ under $q_{\theta}^i$.
\item Let $\sB_i$ be the $B$ elements of $\tilde{\sB}_i$ with highest mass.
\end{itemize}
The similarity can be made precise: beam search is a degenerate 
instance of our procedure. Given $\sB_i$, let 
$\sC_i = \{\{b\} \mid b \in \sB_i\} \cup \{\sY_{1:i}\}$. Then $\sC_i$ consists 
of singleton sets for each element of $\sB_i$, together with $\sY_{1:i}$ in order 
to ensure coverage. To get back to beam search (which doesn't have coverage), we 
add an additional feature to our model: $\bI[c_i = \sY_{1:i}]$. We set the 
weight of this feature to $-\infty$, assigning zero mass to 
everything outside of $\sB_i$. 

Given any algorithm based on beam search, we can improve it simply by 
allowing the weight on this additional feature to be learned from data. 
This can help with the precision ceiling issue by allowing us to reason 
about when beam search is likely to have made a search error.

\subsection{Featurizations}
We end this section with a recipe for choosing features 
$\phi_i(c_{i-1}, y_i)$. We focus on $n$-gram and alignment features, 
which are what we use in our experiments.

\paraii{$n$-gram features.} We consider $n$th-order Markov chains 
over text, typically featurized by $(n+1)$-grams:
\[ \phi_i(y_{1:i-1}, y_i) = \big(\bI[y_{i-n:i} = \hat{y}]\big)_{\hat{y} \in \sY_{i-n:i}}. \]
To extend this to our setting, define $\overline{\sY}_i = \sY_i \cup \{\star\}$ 
and $\overline{\sY}_{1:i} = \prod_{j=1}^i \overline{\sY}_i$. 
We can identify each pair 
$(c_{i-1}, y_i)$ with a sequence $s = \sigma(c_{i-1},y_i) \in \overline{\sY}_i$ 
in the same way as before: 
in each position $j \leq i$ where $y_j$ is determined by 
$(c_{i-1}, y_i)$, $s_j = y_j$; otherwise, $s_j = \star$. We then define 
our $n$-gram model on the extended space $\overline{\sY}_{i-n:i}$:
\[ \phi_i(c_{i-1}, y_i) = \big(\bI[\sigma(c_{i-1}, y_i) = \hat{y}]\big)_{\hat{y} \in \overline{\sY}_{i-n:i}}. \]

\paraii{Alignments.} 
In the speech task from Section~\ref{sec:tasks}, we have an input 
$x_{1:L'}$ and output $y_{1:L}$, where $x$ and $y$ have different lengths and need to be 
aligned. To capture this, we add an \emph{alignment} $z$ to the model, such as the one below:
{\setlength{\tabcolsep}{0.15pc}
\begin{center}
\begin{tabular}{c|l|l|l|l|l|l|l|l|l|l|l|l|l}
aligned input  & h\# & y & ae  &    & ax &   & s &    & w & ih & r & dcl d & h\# \\
aligned output &     & y & eah & \_ & it & ' & s & \_ & w & ei & r & d     &
\end{tabular}
\end{center}}
We represent $z$ as a bipartite graph between $\{1,\ldots,L\}$ and $\{1,\ldots,L'\}$ with no 
crossing edges, and where every node has degree at least one. 
The non-crossing condition allows 
one phoneme to align to multiple characters, or one 
character to align to multiple phonemes, but not both.
Our goal is to model $p_{\theta}(y,z \mid x)$.

To featurize alignment models, we place $n$-gram features on the output $y_i$, as well as 
on every group of $n$ consecutive edges. In addition, we augment the context $c_{i}$ to 
keep track of what $y_i$ most recently aligned to 
(so that we can ensure the alignment is monotonic). 
We also maintain the $B$ best contexts at position $i$ separately for each of the $L'$ possible 
values of $z_i$; this modification to the \method heuristic encourages even 
coverage of the space of alignments.

%% file: precision.stex
\section{Generating High Precision Predictions}
\label{sec:precision}

Recall that one symptom stemming from a lack of coverage is poor estimates of
uncertainty and the inability to generate high precision predictions.
In this section, we show that the coverage offered by \method mitigates this
issue compared to beam search.

Specifically, we are interested in whether an algorithm can find a large subset of test examples 
that it can classify with high ($\approx 99\%$) accuracy. 
Formally, assume a method outputs a prediction $y$ with confidence $p \in [0,1]$ for each example. 
We sort the examples by confidence, and see what fraction $R$ of examples 
we can answer before our accuracy drops below a given threshold $P$. 
In this case, $P$ is the \emph{precision} and $R$ is the \emph{recall}.

Having good recall at high levels of precision (e.g., $P=0.99$) is useful in applications 
where we need to pass on predictions below the precision threshold for a human to verify,
but where we would still like 
to classify as many examples as possible automatically.

We ran an experiment on the word recognition dataset described 
in Section~\ref{sec:tasks}. 
We used a $4$-gram model, training both beam search (with a beam size of $10$) and 
\method (with $10$ contexts per position, not counting $\sY_{1:i}$). In addition, we used beam search with a beam size of 
$200$ to simulate almost-exact inference. To train the models,
we maximized the approximate log-likelihood
using AdaGrad \citep{duchi2011adaptive} 
with a step size $\eta$ of $0.2$ and $\delta = 10^{-4}$.

\begin{figure}
\begin{center}
\includegraphics[width=0.9\columnwidth]{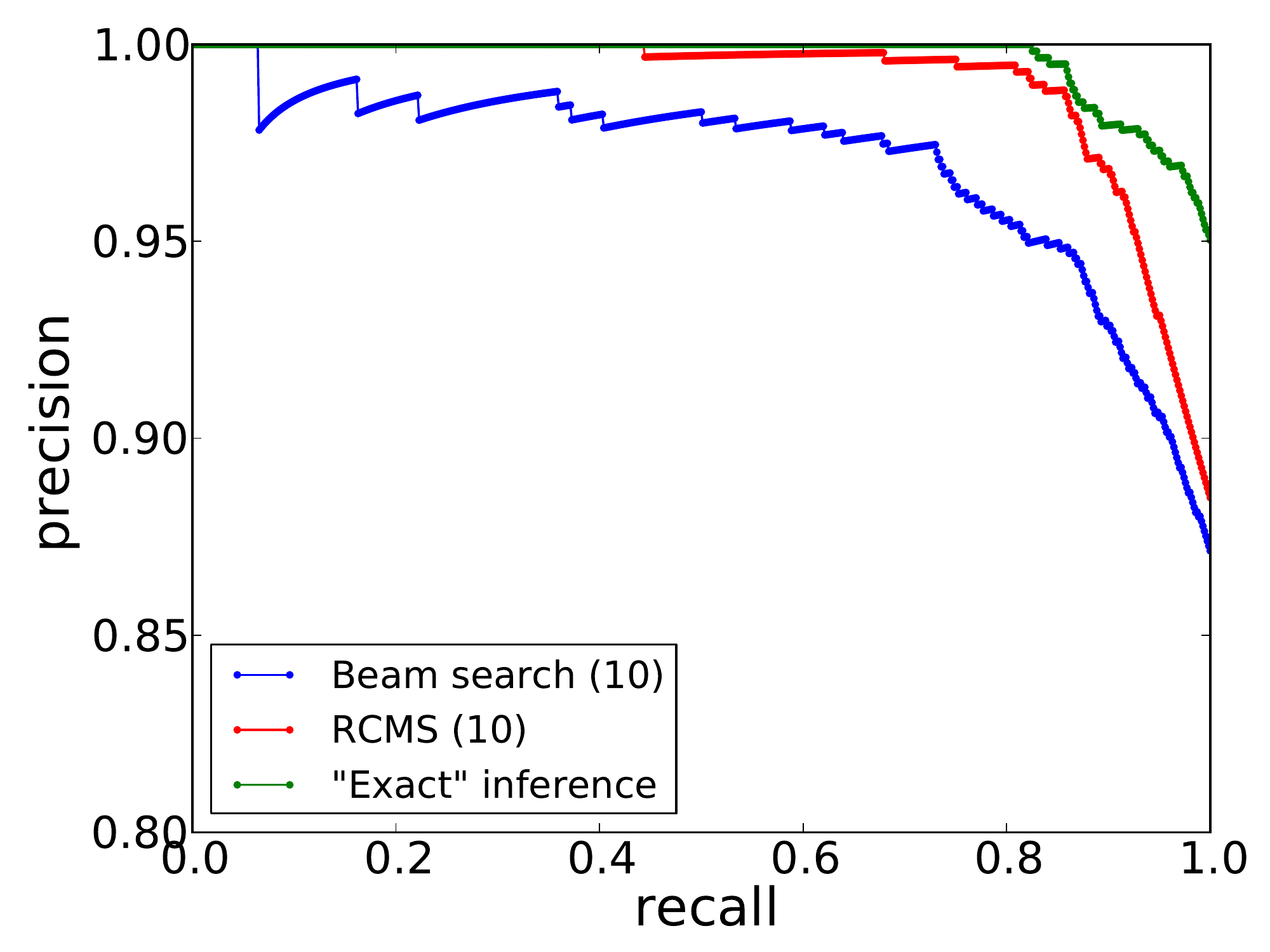}
\end{center}
\vspace{-0.07in}
\caption{On word recognition,
precision-recall curve of
beam search with beam size $10$,
\method with $10$ contexts per position, and 
almost-exact inference simulated by beam search with a beam of size $200$.
Beam search makes errors even on its most confident predictions, 
while \method is able to separate out a large number of nearly error-free 
predictions.}
\label{fig:precision}
\vspace{-0.12in}
\end{figure}

The precision-recall curve for each method is plotted in Figure~\ref{fig:precision}; 
confidence is the probability the model assigns to the predicted output.
Note that while beam search and \method achieve similar accuracies (precision at $R = 1$)
on the full test set ($87.1\%$ and $88.5\%$, respectively),
\method is much better at separating out examples 
that are likely to be correct. The flat region in the precision-recall curve for beam 
search means that the model confidence and actual error probability are
unrelated across that region.

As a result, there is a \emph{precision ceiling},
where it is simply impossible to obtain high precision at any reasonable level of
recall.
To quantify this effect, note that the recall at $99\%$ precision for 
beam search is only $16\%$, while for \method it is $82\%$. For comparison, the recall 
for exact inference is only $4\%$ higher ($86\%$).
Therefore, \method is nearly as effective as exact inference on this metric
while requiring substantially fewer computational resources.

%% file: accuracy.stex
\section{Learning with Indirect Supervision}
\label{sec:accuracy}

The second symptom of lack of coverage is the inability to learn from indirect supervision.
In this setting, we have an exponential family model
$p_{\theta}(y, z \mid x) \propto \exp(\theta^{\top}\phi(x, y,z))$,
where $x$ and $y$ are observed during training but $z$ is unobserved.
The gradient of the (marginal) log-likelihood is:
\begin{align}
\nabla \log p_{\theta}(y \mid x) &= \bE_{\hat{z} \sim p_{\theta}(z \mid x, y)}\left[\phi(x, y,\hat{z})\right]\\
\notag
 &\phantom{==} - \bE_{\hat{y},\hat{z} \sim p_{\theta}(y,z \mid x)}\left[\phi(x, \hat{y},\hat{z})\right],
\end{align}
which is the difference between the expected features
with respect to the \emph{target} distribution $p_\theta(z \mid x, y)$ and
the \emph{model} distribution $p_\theta(y, z \mid x)$.
In the fully supervised case, where we observe $z$, the target term is simply $\phi(x, y, z)$,
which provides a clear training signal without any inference.
With indirect supervision, even obtaining a training signal
requires inference with respect to $p_\theta(z \mid x, y)$,
which is generally intractable.

In the context of beam search,
there are several strategies to inferring $z$ for computing gradients:
\begin{itemize}[itemsep=2pt,topsep=0pt,parsep=0pt,partopsep=0pt,leftmargin=12pt]
\item \textbf{Select-by-model:} select beams based on $q_{\theta}^i(z\! \mid\! x)$, 
      then re-weight at the end by $p_{\theta}(y\! \mid\! z, x)$.
      This only works if the weights
      are high for at least some ``easy'' examples, from which learning
      can then bootstrap.
\item \textbf{Select-by-target:} select beams based on $q_{\theta}^i(z\! \mid\! x,y)$.
      Since $y$ is not available at test time,
      parameters $\theta$ learned conditioned on $y$ do not generalize well.
\item \textbf{Hybrid:} take the union of beams based on both the model and target distributions.
\item \textbf{Forced decoding \citep{gorman2011prosodylab}:} 
      first train a simple model for which exact inference is 
      tractable to infer the most likely $z$, conditioned on $x$ and $y$.  Then simply fix $z$;
      this becomes a fully-supervised problem.
\end{itemize}

To understand the behavior of these methods, we used them all to train a model on the 
speech recognition dataset from Section~\ref{sec:tasks}. The model places $5$-gram indicator features 
on the output as well as on the alignments.
We trained using AdaGrad with step size $\eta = 0.2$ 
and $\delta = 10^{-4}$. For each method, we set the beam size to $20$.
For forced decoding, we used a bigram model with exact inference to 
impute $z$ at the beginning.

\begin{figure*}
\centering
\subfigure{\label{fig:accuracy-speech-beam}\includegraphics[width=0.32\textwidth]{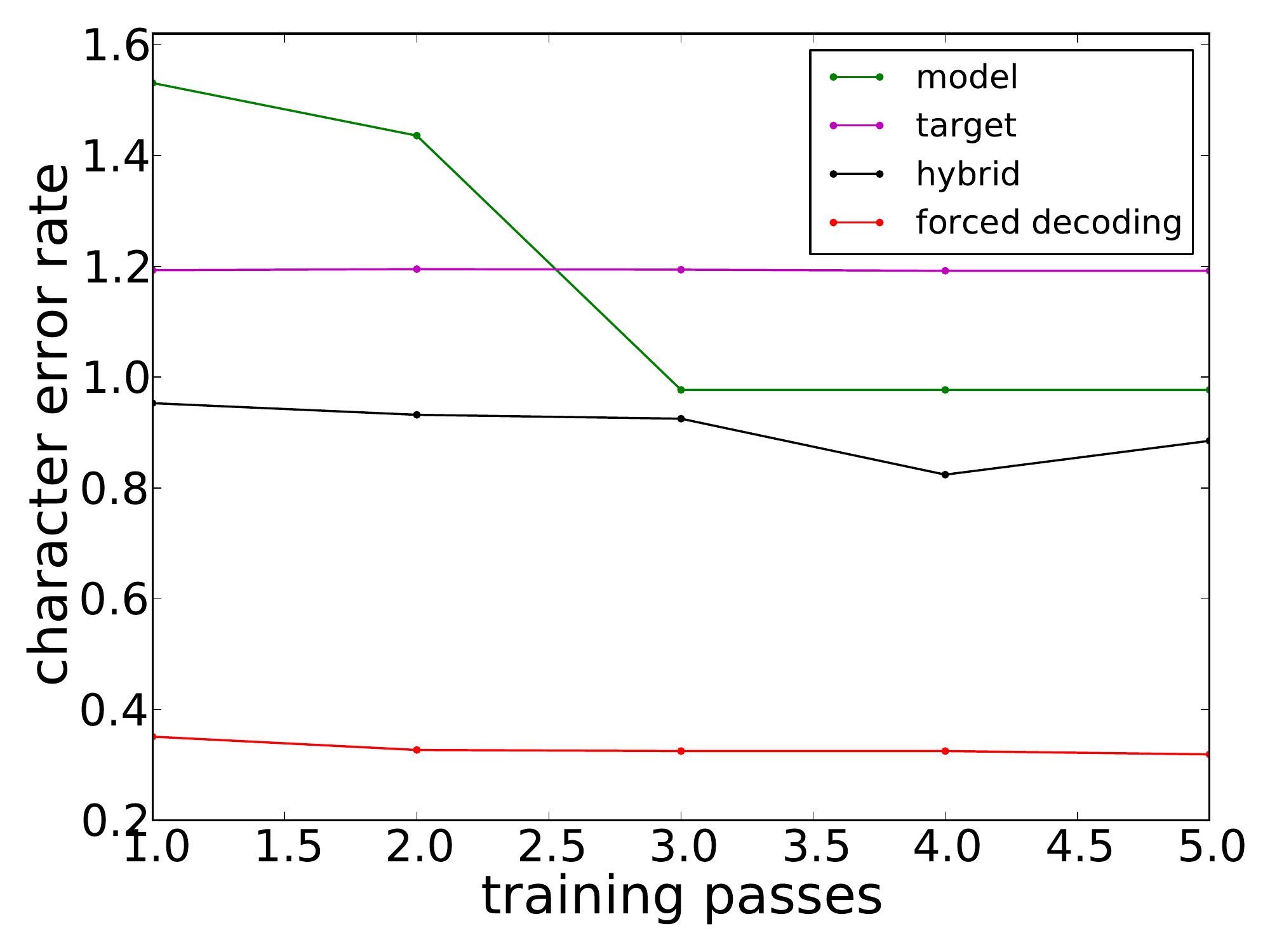}}
\subfigure{\label{fig:accuracy-speech-char}\includegraphics[width=0.32\textwidth]{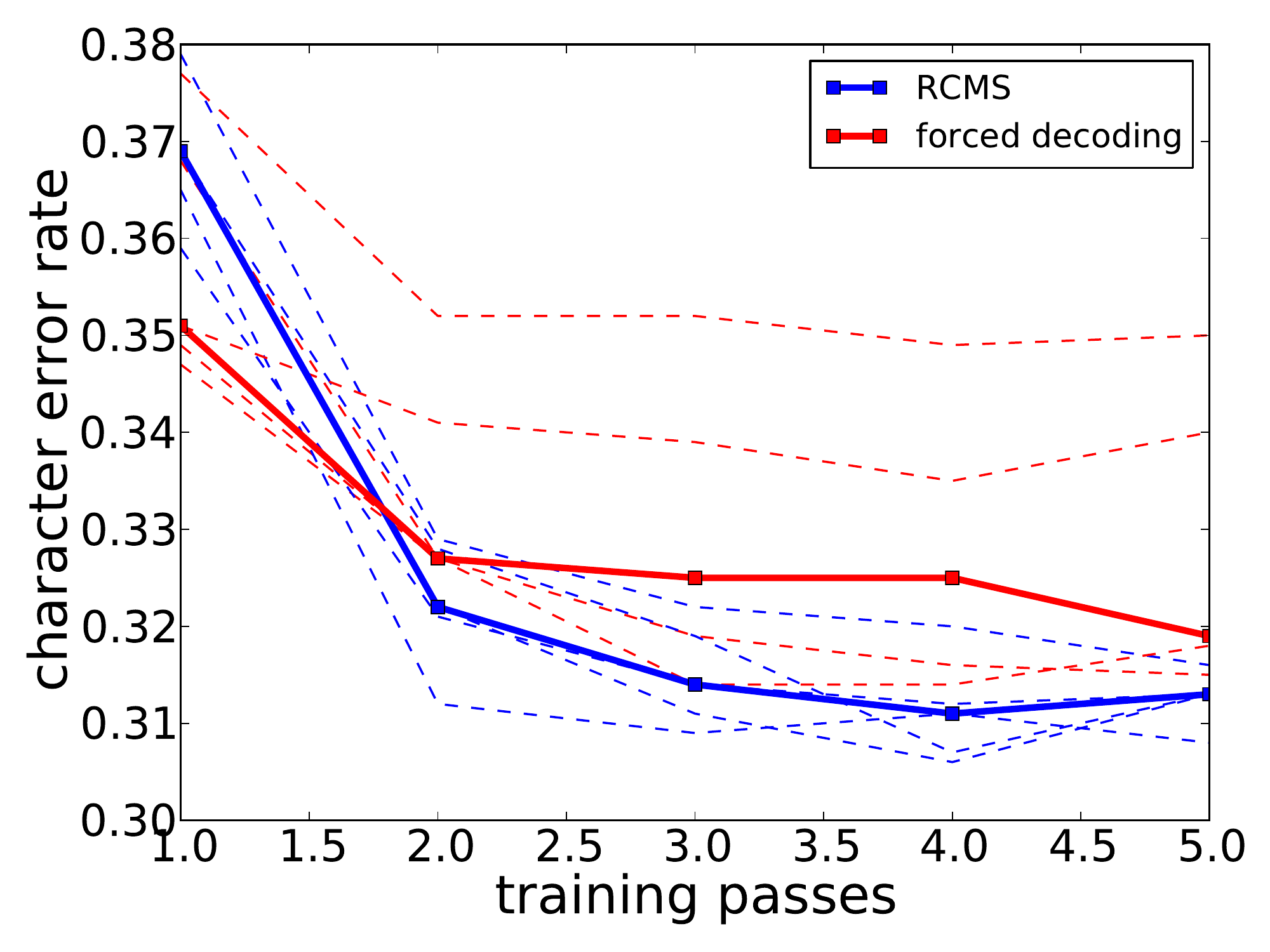}}
\subfigure{\label{fig:accuracy-speech-exact}\includegraphics[width=0.32\textwidth]{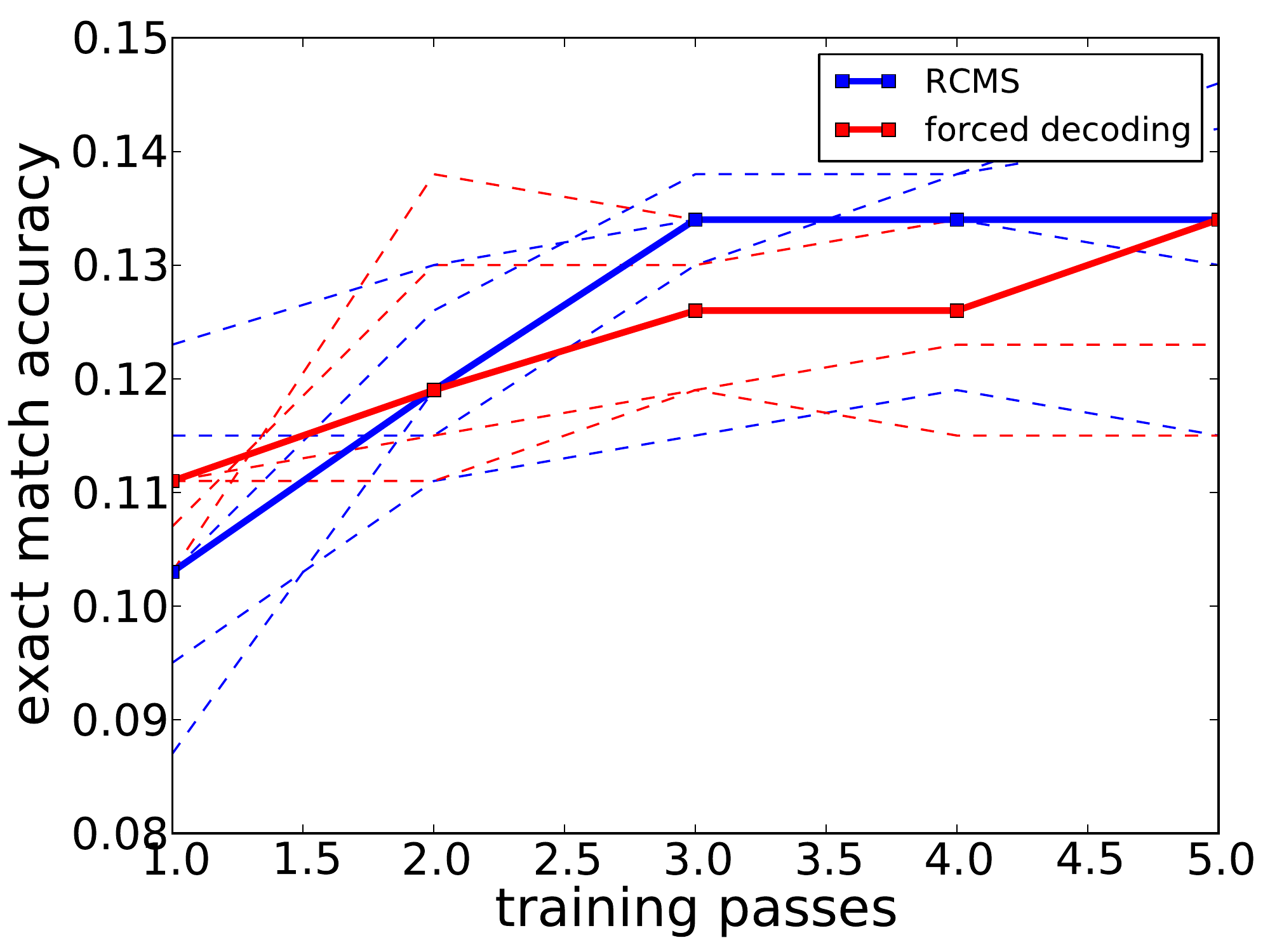}}
\vspace{-0.07in}
\caption{Left: character error rate (CER) of all beam search-based methods on the speech task, 
for $5$ passes of the training data; note that an empty output always has a CER of $1.0$. 
Middle: CER of forced decoding and \method over 
$5$ random permutations of the data; the solid line is the median. Right: exact-match accuracy 
over the same $5$ permutations.}
\label{fig:accuracy-speech}
\end{figure*}
The results are shown in Figure~\ref{fig:accuracy-speech-beam}. Select-by-model doesn't learn 
at all: it only finds valid alignments for $2$ out of the $746$ training examples; for the rest, 
$p_{\theta}(y \mid z,x)$ is zero for all alignments considered, thus
providing no signal for learning.
Select-by-target quickly reaches high training accuracy, but generalizes extremely 
poorly because it doesn't learn to keep the right answer on the beam.
The hybrid approach does better but still not very well. 
The only method that learns effectively is forced decoding.

While forced decoding works well, it relies on the idea that a simple model 
can effectively determine $z$ given access to $x$ and $y$. This will not always 
be the case, so we would like methods that work well even without such a model.
Reified context models provide a natural way of doing this: 
we simply compute $p_{\theta}(z \mid x,y)$ under the contexts selected by \methodp, 
and perform learning updates in the natural way. 

To test \methodp, we trained it in the same way 
using $20$ contexts per position.
Without any need for an initialization scheme, 
we obtain a model whose test accuracy is better than that of forced 
decoding (see Figures~\ref{fig:accuracy-speech-char},\ref{fig:accuracy-speech-exact}). 

\paragraph{Decipherment: Unsupervised Learning.}

We now turn our attention to an unsupervised problem: the decipherment 
task from Section~\ref{sec:tasks}. 
We model decipherment as a hidden Markov model (HMM): the 
hidden plain text evolves according to an $n$-th order
Markov chain, and the cipher text 
is emitted based on a deterministic but unknown 1:1 substitution cipher 
\citep{ravi2009attacking}.

All the methods we described for speech recognition break down in the 
absence of any supervision except select-by-model. We therefore 
compare only three methods: select-by-model (beam search), \methodp, and exact inference. 
We trained a $1$st-order (bigram) HMM using all 3 methods, and a $2$nd-order (trigram)
HMM using only beam search and \methodp,
as exact inference was too slow (the vocabulary size is $500$).
We used the given plain text to 
learn the transition probabilities, using absolute 
discounting \citep{ney1994structuring} for smoothing. Then, we used EM to learn the transition 
probabilities; we used Laplace smoothing for these updates.

\begin{figure*}[tb!]
\begin{center}
\includegraphics[width=0.45\textwidth]{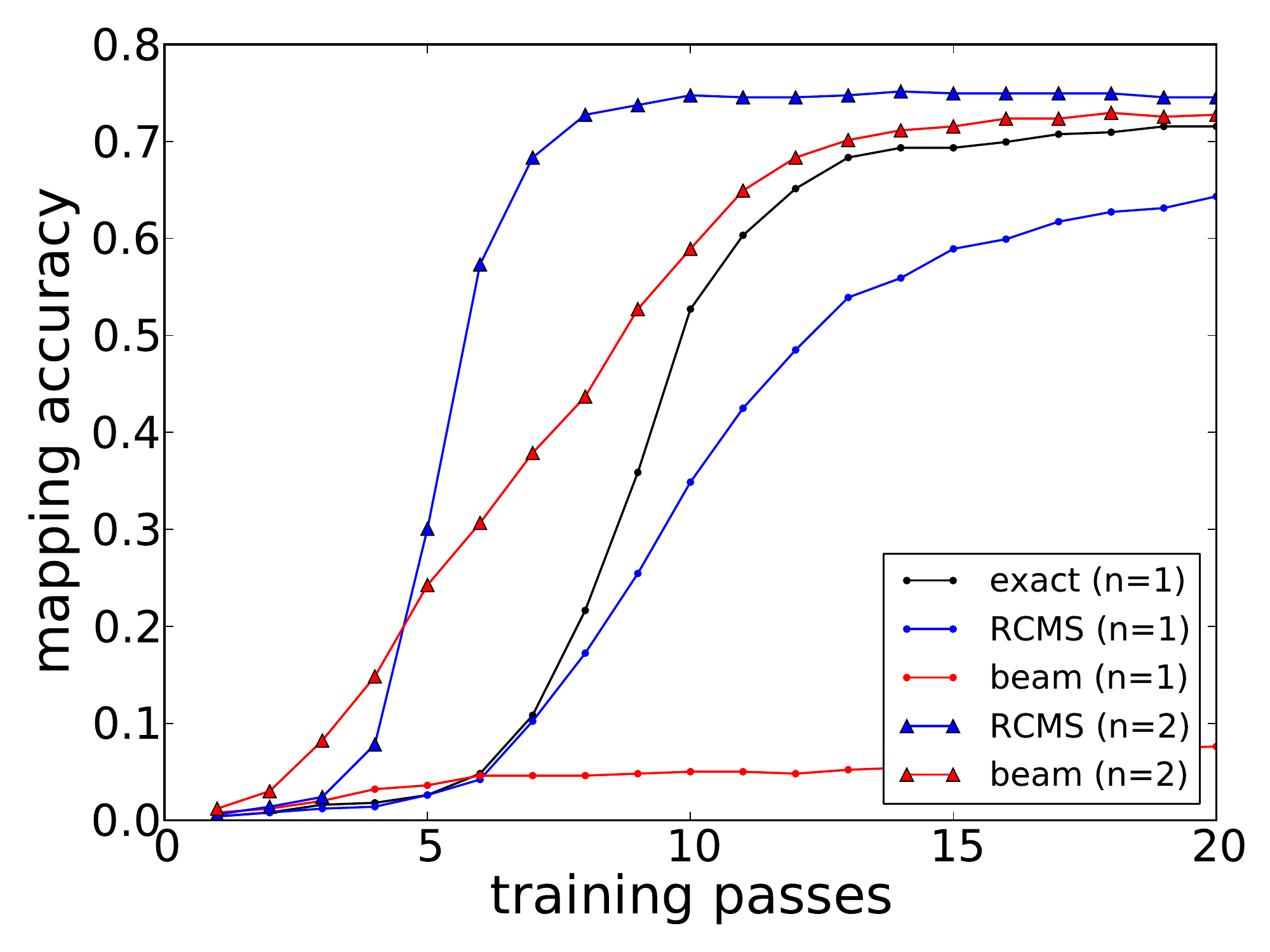}
\includegraphics[width=0.45\textwidth]{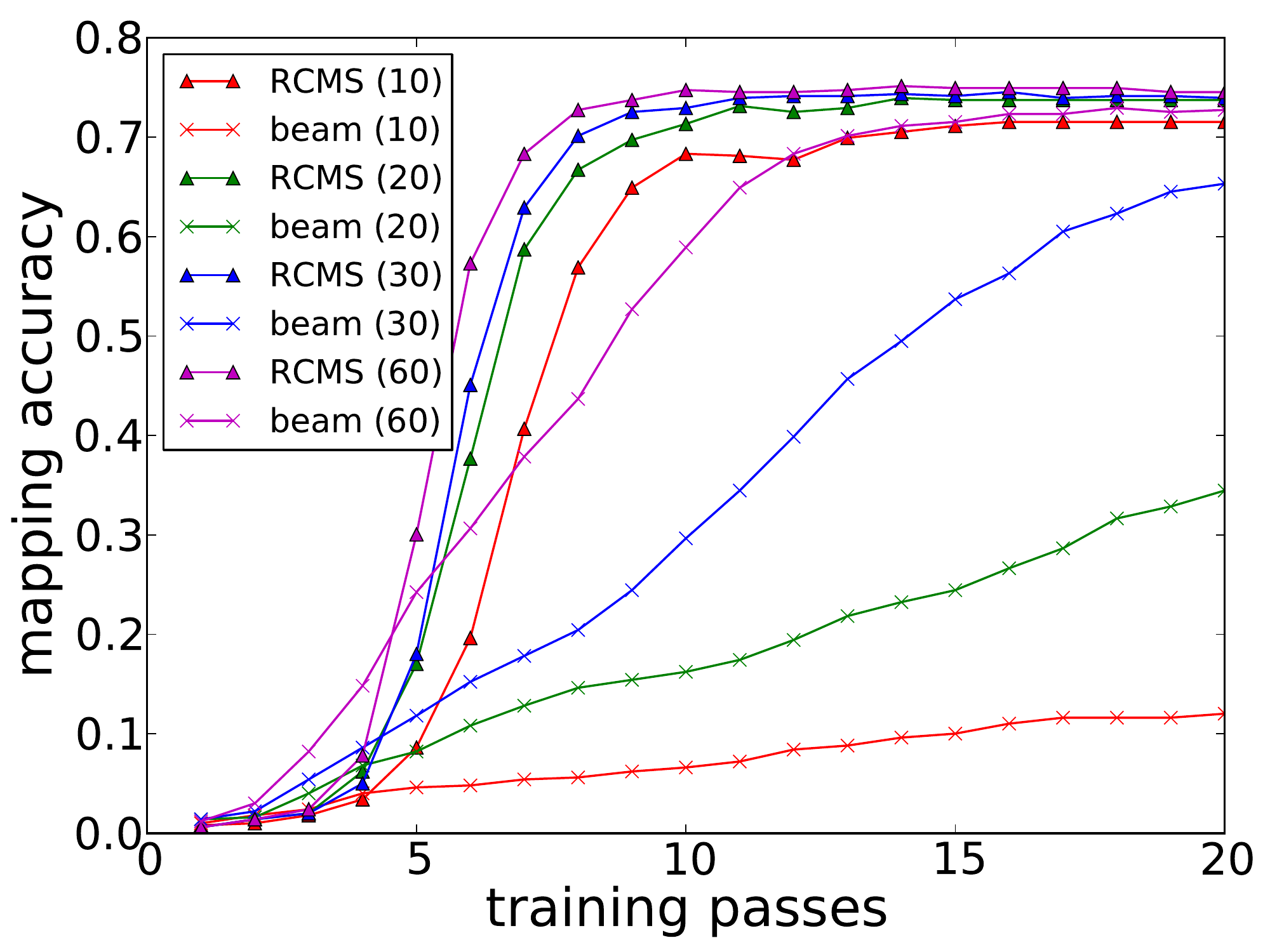}
\end{center}
\vspace{-0.15in}
\caption{Results on the decipherment task. Left: accuracy for a fixed 
beam/context size as the model order varies; approximate inference with a $2$nd-order HMM
using \method outperforms both beam search in the same model
and exact inference in a simpler model.
Right: effect of beam/context size on accuracy for the $2$nd-order HMM.
\method is much more robust to changes in beam/context size.}
\label{fig:accuracy-decipherment}
\end{figure*}
The results are shown in Figure~\ref{fig:accuracy-decipherment}. We measured 
performance by mapping accuracy: the fraction of unique symbols that are correctly mapped 
\citep{nuhn2013beamdecipher}.
First, we compared the overall accuracy of all methods, 
setting the beam size and context size both to $60$. We see that all 
$2$nd-order models outperform all $1$st-order models, and that beam search 
barely learns at all for the $1$st-order model. 

Restricting attention to $2$nd-order models, we measure the effect of beam size 
and context size on accuracy, plotting learning curves for sizes of $10$, $20$, $30$, and $60$. 
In all cases, \method learns more quickly and converges 
to a more accurate solution than beam search. The shapes of the learning curves are also 
different: \method learns quickly after a few initial iterations, 
while beam search slowly accrues information at a roughly constant rate over time.

%% file: context.stex
\section{Refinement of Contexts During Training}
\label{sec:context}

When learning with indirect supervision and approximate inference,
one intuition is that we can ``bootstrap'' 
by first learning from easy examples, and then using the information gained from these 
examples to make better inferences about the remaining ones \citep{liang11dcs}. 
However, this can fail if there are insufficiently many easy examples (as in the speech task), 
if the examples are hard to identify, or if they differ statistically
from the remaining examples. 

We think of the above as ``vertical bootstrapping'':
using the full model on an increasing number of examples.
\method instead performs ``horizontal 
bootstrapping'': for each example, it
selects a model (via the context sets) based on
the information available.
As training progresses, we expect these 
contexts to become increasingly fine as our parameters improve.

To measure this quantitatively, we define the 
\emph{length} of a context $c_{i-1}$ to be the number of positions of 
$y_{1:i-1}$ that can be determined from $c_{i-1}$ (number of non-$\star$'s).
We plot the average 
length (weighted by mass under $q_{\theta}^i$) as training progresses.
The averages are 
updated every $50$ and $100$ training examples respectively 
for word and speech recognition. 
For decipherment, they are computed 
once for each full pass over the training data (since EM only updates the parameters once 
per pass).

\begin{figure*}[tb!]
\begin{center}
\includegraphics[width=0.31\textwidth]{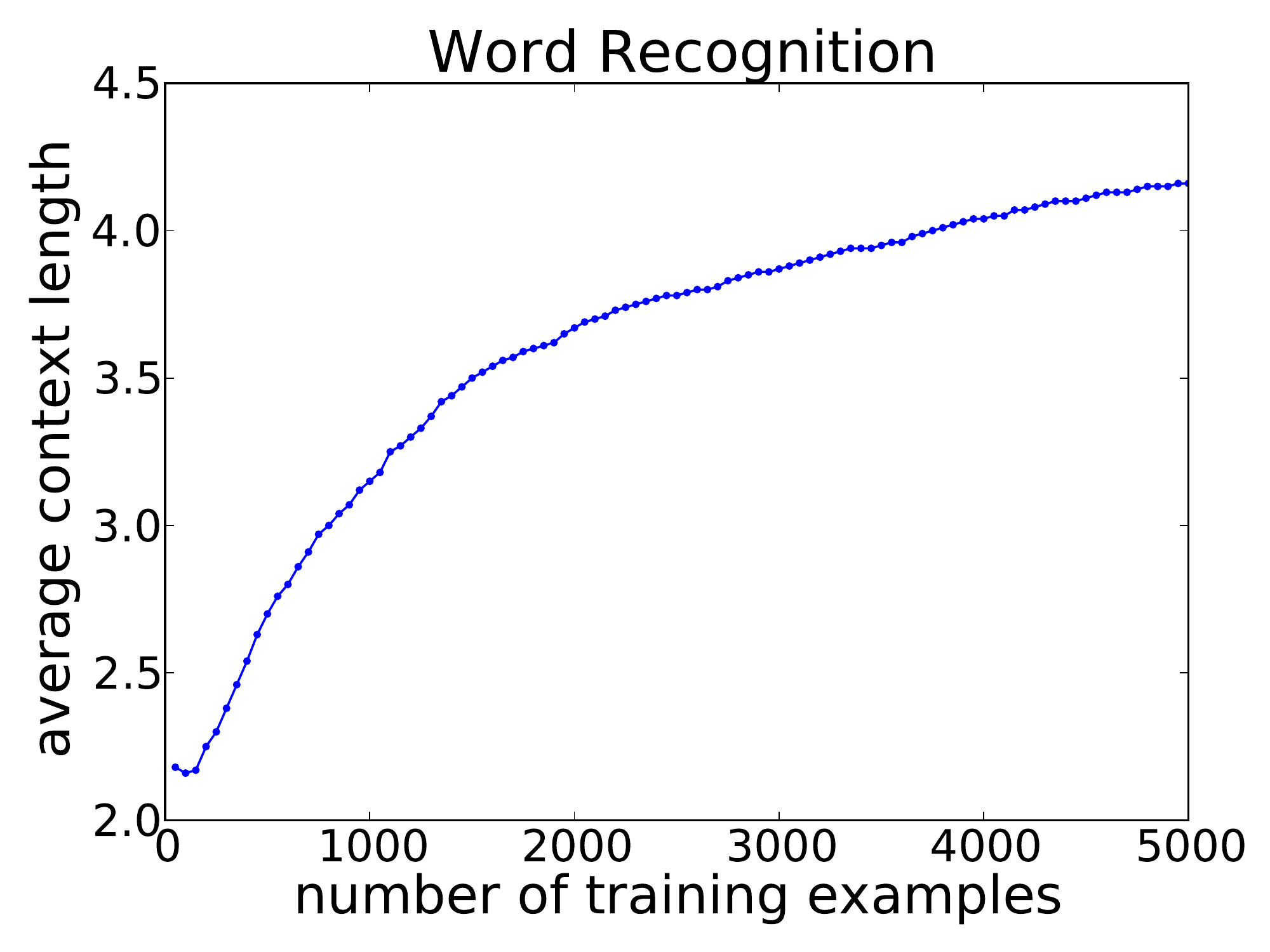}
\includegraphics[width=0.31\textwidth]{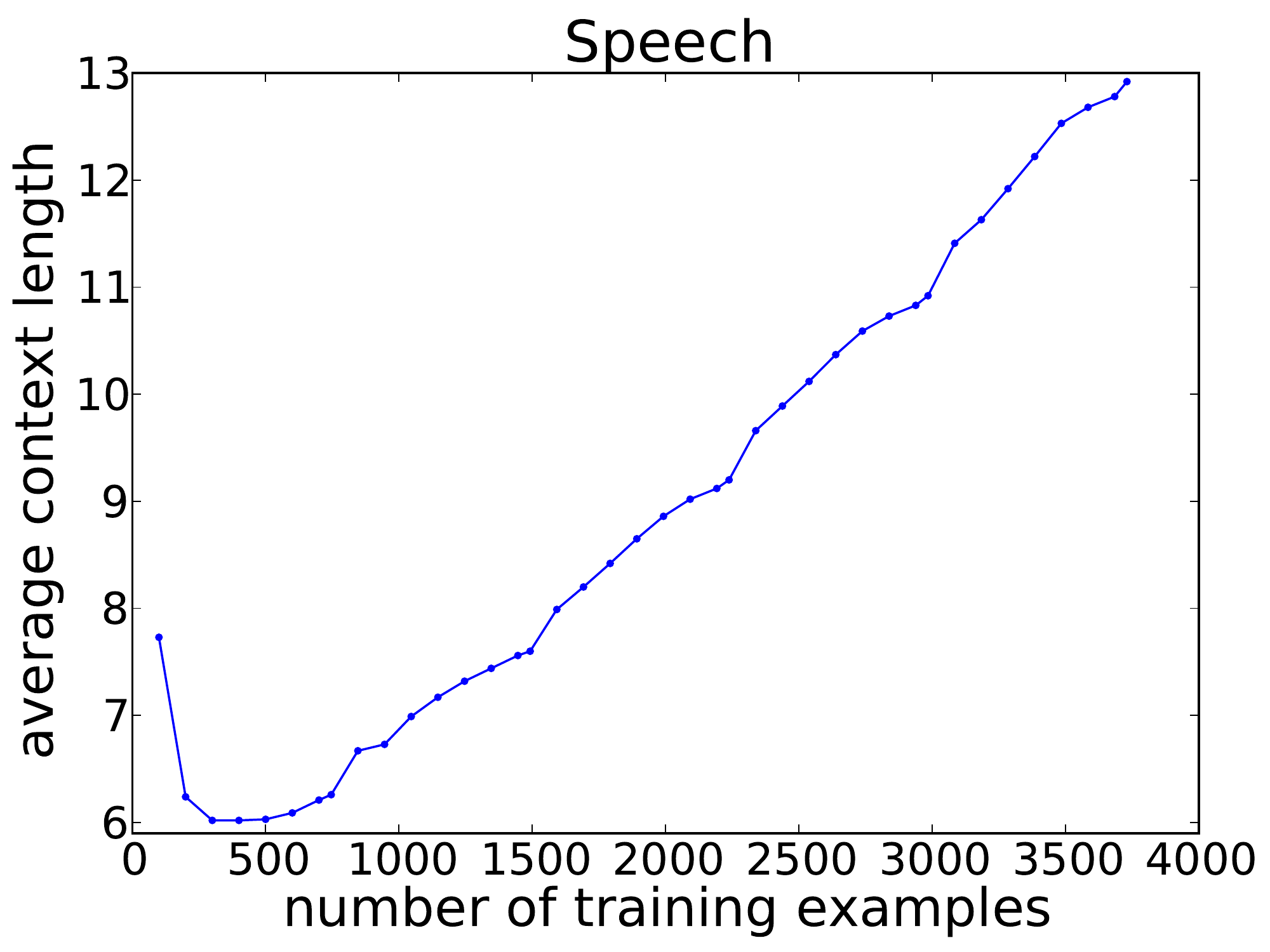}
\includegraphics[width=0.31\textwidth]{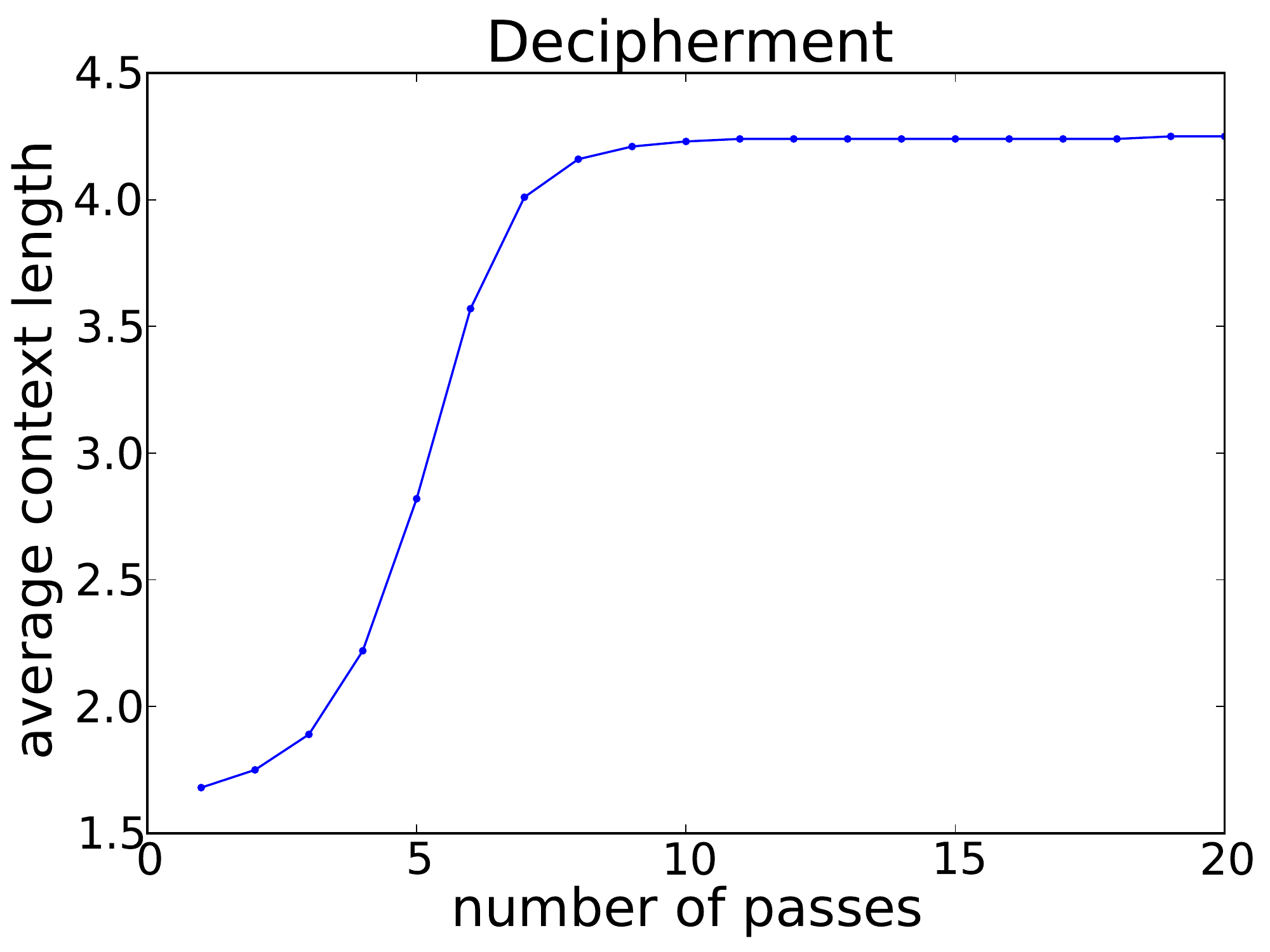}
\end{center}
\vspace{-0.07in}
\caption{Average context length vs. number of learning updates for 
the word recognition, speech, and decipherment tasks. For word and 
speech recognition we take a cumulative average (to reduce noise).}
\label{fig:context}
\vspace{-0.15in}
\end{figure*}
Figure~\ref{fig:context} shows that the broad trend is an increase
in the context length over time.
For both the word and speech tasks, there is an initial overshoot at the beginning that is not 
present in the decipherment task; this is because the word and speech tasks 
are trained with stochastic gradient methods, which often overshoot and then 
correct in parameter space, while for decipherment we use the more stable EM 
algorithm.

Since we start by using coarse contexts and move to finer contexts by 
the end of training, \method can be thought of as a coarse-to-fine 
training procedure \citep{petrov2011coarse}. However, instead of using a pre-defined, discrete 
set of models for initialization, we organically adapt the amount of context 
on a per-example basis.

%% file: discussion.stex
\section{Related work} 
\citet{kulesza2007structured} first study the interaction between 
approximate inference and learning, showing that even in the 
fully supervised case approximate inference can be seriously 
detrimental; \citet{finley2008training} show that approximate 
inference
algorithms which over-generate possible outputs interact best with learning; 
this further supports the need for coverage when learning. 

Four major approaches have been taken to address the problem 
of learning with inexact inference. The first 
modifies the learning updates to account for the inference procedure, 
as in the max-violation perceptron and related algorithms 
\citep{huang2012structured,zhang2013online,yu2013max};
reinforcement learning approaches to inference 
\citep{daume2009search,shi2015learning} also fit into this category.
Another approach modifies the inference algorithm to obtain 
better coverage, as in coarse-to-fine inference \citep{petrov06latent,weiss2010sidestepping}, 
where simple models are used to direct the focus of more complex models.
\citet{pal2006sparse} encourage coverage for beam search by 
adaptively increasing the beam size.
A third approach is to use inference procedures with certificates of 
optimality, based on either duality gaps from convex programs \citep{sontag2010approximate} 
or variational bounds \citep{xing2002generalized,wainwright2005new}.

Finally, another way of sidestepping the problems of 
approximate inference is to learn a model that is already 
tractable. While classical tractable model families based 
on low treewidth are often 
insufficiently expressive, more modern families have shown promise; 
for instance, sum-product networks \citep{poon2011sum} can express models with 
high treewidth while still being tractable, and have achieved 
state-of-the-art results for some tasks. Other work 
includes exchangeable variable models \citep{niepert2014exchangeable} 
and mean-field networks \citep{li2014mean}.

Our method \method also attempts to define tractable model families,
in our case, via a parsimonious choice of latent context variables,
even though the actual distribution over $y_{1:L}$ may have arbitrarily 
high treewidth. We adaptively choose the model structure for 
each example at ``run-time'', which distinguishes our approach from 
the aforementioned methods,
though sum-product networks have some capacity for expressing adaptivity implicitly.
We believe that such per-example adaptivity is important for obtaining
good performance on challenging structured prediction tasks.

Certain smoothing techniques in natural language processing also 
interpolate between contexts of different order, such as absolute 
discounting \citep{ney1994structuring} and Kneser-Ney 
smoothing \cite{kneser1995improved}. However, in such cases 
\emph{all} observed contexts are used in the model; to get 
the same tractability gains as we do, it would be necessary to 
adaptively sparsify the model for each example at run-time.
Some Bayesian nonparametric approaches such as infinite contingent Bayesian networks 
\citep{milch2005approximate} and hierarchical Pitman-Yor 
processes \citep{teh2006hierarchical,wood2009stochastic} also reason 
about contexts; again, such models do not lead to tractable inference.

\section{Discussion}
\label{sec:discussion}

We have presented a new framework, \emph{reified context models}, that reifies 
context as a random variable, thereby defining a family of expressive but tractable 
probability distributions. By adaptively choosing context sets at run-time, 
our \method method 
uses short contexts in regions of high uncertainty 
and long contexts in regions of low uncertainty, thereby reproducing the 
behavior of coarse-to-fine training methods in a more organic and 
fine-grained manner. In addition, because \method maintains full coverage of 
the space, it is able to break through the precision ceiling faced by 
beam search. Coverage also helps with training under indirect supervision,
since we can better identify settings of latent variables that assign high 
likelihood to the data.

At a high level, our method provides a framework for structuring 
inference in terms of the contexts it considers; because the contexts are 
reified in the model, we can also support queries about how much 
probability mass lies in each context. These two properties together open up 
intriguing possibilities. For instance, one could imagine a multi-pass approach 
to inference where the first pass uses small context sets for each 
location, and later passes add additional contexts at locations where there is 
high uncertainty. By adaptively adding context only when it is needed, 
we could speed up inference by a potentially large amount.

Another direction of research is to extend our construction beyond 
a single left-to-right ordering. In principle, we can consider 
any collection of contexts that induce a graphical model with low treewidth, 
rather than only considering the factorization in \eqref{pic:graphical-model}. 
For problems such as image segmentation where the natural structure is a 
grid rather than a chain, such extensions may be necessary.

Finally, while we currently learn how much weight to assign to each context, 
we could go one step further and learn which contexts to propose and include 
in the context sets $\sC_i$ (rather than 
relying on a fixed procedure as in the \method algorithm). Ideally, one could 
specify a large number of possible strategies for building context sets, and the 
best strategy to use for a given example would be learned from data. 
This would move us one step closer to being able to employ arbitrarily
expressive models with the assurance of an automatic inference procedure
that can take advantage of the expressivity in a reliable manner.

%% file: arxiv.bbl
\begin{thebibliography}{39}
\providecommand{\natexlab}[1]{#1}
\providecommand{\url}[1]{\texttt{#1}}
\expandafter\ifx\csname urlstyle\endcsname\relax
  \providecommand{\doi}[1]{doi: #1}\else
  \providecommand{\doi}{doi: \begingroup \urlstyle{rm}\Url}\fi

\bibitem[Brooks et~al.(2011)Brooks, Gelman, Jones, and
  Meng]{brooks2011handbook}
Brooks, Steve, Gelman, Andrew, Jones, Galin, and Meng, Xiao-Li.
\newblock \emph{Handbook of Markov Chain Monte Carlo}.
\newblock CRC press, 2011.

\bibitem[Capp{\'e} et~al.(2007)Capp{\'e}, Godsill, and Moulines]{smc}
Capp{\'e}, Olivier, Godsill, Simon~J, and Moulines, Eric.
\newblock An overview of existing methods and recent advances in sequential
  {M}onte {C}arlo.
\newblock \emph{Proceedings of the IEEE}, 95\penalty0 (5):\penalty0 899--924,
  2007.

\bibitem[Daum{\'e}~{III} et~al.(2009)Daum{\'e}~{III}, Langford, and
  Marcu]{daume2009search}
Daum{\'e}~{III}, Hal, Langford, John, and Marcu, Daniel.
\newblock Search-based structured prediction.
\newblock \emph{Machine learning}, 75\penalty0 (3):\penalty0 297--325, 2009.

\bibitem[Doucet \& Johansen(2011)Doucet and Johansen]{particle-filter}
Doucet, Arnaud and Johansen, Adam~M.
\newblock A tutorial on particle filtering and smoothing: Fifteen years later.
\newblock In \emph{Oxford Handbook of Nonlinear Filtering}. Citeseer, 2011.

\bibitem[Duchi et~al.(2010)Duchi, Hazan, and Singer]{duchi2011adaptive}
Duchi, J., Hazan, E., and Singer, Y.
\newblock Adaptive subgradient methods for online learning and stochastic
  optimization.
\newblock In \emph{Conference on Learning Theory (COLT)}, 2010.

\bibitem[Finley \& Joachims(2008)Finley and Joachims]{finley2008training}
Finley, Thomas and Joachims, Thorsten.
\newblock Training structural svms when exact inference is intractable.
\newblock In \emph{Proceedings of the 25th international conference on Machine
  learning}, pp.\  304--311. ACM, 2008.

\bibitem[Gorman et~al.(2011)Gorman, Howell, and Wagner]{gorman2011prosodylab}
Gorman, Kyle, Howell, Jonathan, and Wagner, Michael.
\newblock Prosodylab-aligner: A tool for forced alignment of laboratory speech.
\newblock \emph{Canadian Acoustics}, 39\penalty0 (3):\penalty0 192--193, 2011.

\bibitem[Graff \& Cieri(2003)Graff and Cieri]{graff2003}
Graff, David and Cieri, Christopher.
\newblock \emph{English Gigaword LDC2003T05}.
\newblock Philadelphia: Linguistic Data Consortium, 2003.
\newblock Web Download.

\bibitem[Greenberg et~al.(1996)Greenberg, Hollenback, and
  Ellis]{greenberg1996insights}
Greenberg, Steven, Hollenback, Joy, and Ellis, Dan.
\newblock Insights into spoken language gleaned from phonetic transcription of
  the switchboard corpus.
\newblock In \emph{Proceedings of the International Conference on Spoken
  Language Processing}, 1996.

\bibitem[Huang et~al.(2012)Huang, Fayong, and Guo]{huang2012structured}
Huang, Liang, Fayong, Suphan, and Guo, Yang.
\newblock Structured perceptron with inexact search.
\newblock In \emph{Proceedings of the 2012 Conference of the North American
  Chapter of the Association for Computational Linguistics: Human Language
  Technologies}, pp.\  142--151. Association for Computational Linguistics,
  2012.

\bibitem[Kassel(1995)]{kassel1995comparison}
Kassel, Robert~H.
\newblock \emph{A comparison of approaches to on-line handwritten character
  recognition}.
\newblock PhD thesis, Massachusetts Institute of Technology, 1995.

\bibitem[Kneser \& Ney(1995)Kneser and Ney]{kneser1995improved}
Kneser, Reinhard and Ney, Hermann.
\newblock Improved backing-off for m-gram language modeling.
\newblock In \emph{Acoustics, Speech, and Signal Processing, 1995. ICASSP-95.,
  1995 International Conference on}, volume~1, pp.\  181--184. IEEE, 1995.

\bibitem[Koehn et~al.(2003)Koehn, Och, and Marcu]{koehn2003statistical}
Koehn, Philipp, Och, Franz~Josef, and Marcu, Daniel.
\newblock Statistical phrase-based translation.
\newblock In \emph{Proceedings of the 2003 Conference of the North American
  Chapter of the Association for Computational Linguistics on Human Language
  Technology-Volume 1}, pp.\  48--54. Association for Computational
  Linguistics, 2003.

\bibitem[Kulesza \& Pereira(2007)Kulesza and Pereira]{kulesza2007structured}
Kulesza, Alex and Pereira, Fernando.
\newblock Structured learning with approximate inference.
\newblock In \emph{Advances in neural information processing systems}, pp.\
  785--792, 2007.

\bibitem[Li \& Zemel(2014)Li and Zemel]{li2014mean}
Li, Yujia and Zemel, Richard.
\newblock Mean-field networks.
\newblock \emph{arXiv preprint arXiv:1410.5884}, 2014.

\bibitem[Liang et~al.(2011)Liang, Jordan, and Klein]{liang11dcs}
Liang, P., Jordan, M.~I., and Klein, D.
\newblock Learning dependency-based compositional semantics.
\newblock In \emph{Association for Computational Linguistics (ACL)}, pp.\
  590--599, 2011.

\bibitem[Milch et~al.(2005)Milch, Marthi, Sontag, Russell, Ong, and
  Kolobov]{milch2005approximate}
Milch, Brian, Marthi, Bhaskara, Sontag, David, Russell, Stuart, Ong, Daniel~L,
  and Kolobov, Andrey.
\newblock Approximate inference for infinite contingent bayesian networks.
\newblock In \emph{Proc. 10th AISTATS}, pp.\  238--245, 2005.

\bibitem[Ney et~al.(1994)Ney, Essen, and Kneser]{ney1994structuring}
Ney, Hermann, Essen, Ute, and Kneser, Reinhard.
\newblock On structuring probabilistic dependences in stochastic language
  modelling.
\newblock \emph{Computer Speech \& Language}, 8\penalty0 (1):\penalty0 1--38,
  1994.

\bibitem[Niepert \& Domingos(2014)Niepert and
  Domingos]{niepert2014exchangeable}
Niepert, Mathias and Domingos, Pedro.
\newblock Exchangeable variable models.
\newblock \emph{arXiv preprint arXiv:1405.0501}, 2014.

\bibitem[Nuhn \& Ney(2014)Nuhn and Ney]{nuhn2014homophonics}
Nuhn, Malte and Ney, Hermann.
\newblock Improved decipherment of homophonic ciphers.
\newblock In \emph{Conference on Empirical Methods in Natural Language
  Processing}, Doha, Qatar, October 2014.

\bibitem[Nuhn et~al.(2013)Nuhn, Schamper, and Ney]{nuhn2013beamdecipher}
Nuhn, Malte, Schamper, Julian, and Ney, Hermann.
\newblock Beam search for solving substitution ciphers.
\newblock In \emph{Annual Meeting of the Assoc. for Computational Linguistics},
  pp.\  1569--1576, Sofia, Bulgaria, August 2013.
\newblock URL \url{http://aclweb.org/anthology//P/P13/P13-1154.pdf}.

\bibitem[Pal et~al.(2006)Pal, Sutton, and McCallum]{pal2006sparse}
Pal, Chris, Sutton, Charles, and McCallum, Andrew.
\newblock Sparse forward-backward using minimum divergence beams for fast
  training of conditional random fields.
\newblock In \emph{IEEE International Conference on Acoustics, Speech and
  Signal Processing}, volume~5, pp.\  V--V. IEEE, 2006.

\bibitem[Petrov et~al.(2006)Petrov, Barrett, Thibaux, and
  Klein]{petrov06latent}
Petrov, S., Barrett, L., Thibaux, R., and Klein, D.
\newblock Learning accurate, compact, and interpretable tree annotation.
\newblock In \emph{International Conference on Computational Linguistics and
  Association for Computational Linguistics (COLING/ACL)}, pp.\  433--440,
  2006.

\bibitem[Petrov \& Charniak(2011)Petrov and Charniak]{petrov2011coarse}
Petrov, Slav and Charniak, Eugene.
\newblock \emph{Coarse-to-fine natural language processing}.
\newblock Springer Science \& Business Media, 2011.

\bibitem[Poon \& Domingos(2011)Poon and Domingos]{poon2011sum}
Poon, Hoifung and Domingos, Pedro.
\newblock Sum-product networks: A new deep architecture.
\newblock In \emph{Computer Vision Workshops (ICCV Workshops), 2011 IEEE
  International Conference on}, pp.\  689--690. IEEE, 2011.

\bibitem[Ravi \& Knight(2009)Ravi and Knight]{ravi2009attacking}
Ravi, Sujith and Knight, Kevin.
\newblock Attacking letter substitution ciphers with integer programming.
\newblock \emph{Cryptologia}, 33\penalty0 (4):\penalty0 321--334, 2009.
\newblock \doi{10.1080/01611190903030920}.
\newblock URL \url{http://dx.doi.org/10.1080/01611190903030920}.

\bibitem[Recht et~al.(2011)Recht, Re, Wright, and Niu]{recht2011hogwild}
Recht, Benjamin, Re, Christopher, Wright, Stephen, and Niu, Feng.
\newblock Hogwild: A lock-free approach to parallelizing stochastic gradient
  descent.
\newblock In \emph{Advances in Neural Information Processing Systems}, pp.\
  693--701, 2011.

\bibitem[Shi et~al.(2015)Shi, Steinhardt, and Liang]{shi2015learning}
Shi, Tianlin, Steinhardt, Jacob, and Liang, Percy.
\newblock Learning where to sample in structured prediction.
\newblock 2015.

\bibitem[Sontag(2010)]{sontag2010approximate}
Sontag, David~Alexander.
\newblock \emph{Approximate inference in graphical models using LP
  relaxations}.
\newblock PhD thesis, Massachusetts Institute of Technology, 2010.

\bibitem[Steinhardt \& Liang(2014)Steinhardt and
  Liang]{steinhardt2014filtering}
Steinhardt, Jacob and Liang, Percy.
\newblock Filtering with abstract particles.
\newblock In \emph{Proceedings of the 31st International Conference on Machine
  Learning (ICML-14)}, pp.\  727--735, 2014.

\bibitem[Teh(2006)]{teh2006hierarchical}
Teh, Yee~Whye.
\newblock A hierarchical bayesian language model based on pitman-yor processes.
\newblock In \emph{Proceedings of the 21st International Conference on
  Computational Linguistics and the 44th annual meeting of the Association for
  Computational Linguistics}, pp.\  985--992. Association for Computational
  Linguistics, 2006.

\bibitem[Wainwright et~al.(2005)Wainwright, Jaakkola, and
  Willsky]{wainwright2005new}
Wainwright, Martin~J, Jaakkola, Tommi~S, and Willsky, Alan~S.
\newblock A new class of upper bounds on the log partition function.
\newblock \emph{Information Theory, IEEE Transactions on}, 51\penalty0
  (7):\penalty0 2313--2335, 2005.

\bibitem[Weiss et~al.(2010)Weiss, Sapp, and Taskar]{weiss2010sidestepping}
Weiss, David, Sapp, Benjamin, and Taskar, Ben.
\newblock Sidestepping intractable inference with structured ensemble cascades.
\newblock In \emph{Advances in Neural Information Processing Systems}, pp.\
  2415--2423, 2010.

\bibitem[Weiss et~al.(2012)Weiss, Sapp, and Taskar]{weiss2012structured}
Weiss, David, Sapp, Benjamin, and Taskar, Ben.
\newblock Structured prediction cascades.
\newblock \emph{arXiv preprint arXiv:1208.3279}, 2012.

\bibitem[Wood et~al.(2009)Wood, Archambeau, Gasthaus, James, and
  Teh]{wood2009stochastic}
Wood, Frank, Archambeau, C{\'e}dric, Gasthaus, Jan, James, Lancelot, and Teh,
  Yee~Whye.
\newblock A stochastic memoizer for sequence data.
\newblock In \emph{Proceedings of the 26th Annual International Conference on
  Machine Learning}, pp.\  1129--1136. ACM, 2009.

\bibitem[Xing et~al.(2002)Xing, Jordan, and Russell]{xing2002generalized}
Xing, Eric~P, Jordan, Michael~I, and Russell, Stuart.
\newblock A generalized mean field algorithm for variational inference in
  exponential families.
\newblock In \emph{Proceedings of the Nineteenth conference on Uncertainty in
  Artificial Intelligence}, pp.\  583--591. Morgan Kaufmann Publishers Inc.,
  2002.

\bibitem[Yu et~al.(2013)Yu, Huang, Mi, and Zhao]{yu2013max}
Yu, Heng, Huang, Liang, Mi, Haitao, and Zhao, Kai.
\newblock Max-violation perceptron and forced decoding for scalable mt
  training.
\newblock In \emph{EMNLP}, pp.\  1112--1123, 2013.

\bibitem[Zhang et~al.(2013)Zhang, Huang, Zhao, and McDonald]{zhang2013online}
Zhang, Hao, Huang, Liang, Zhao, Kai, and McDonald, Ryan.
\newblock Online learning for inexact hypergraph search.
\newblock In \emph{Proceedings of EMNLP}, 2013.

\bibitem[Zhang et~al.(2014)Zhang, Kalashnikov, and Mehrotra]{zhang2014face}
Zhang, Liyan, Kalashnikov, Dmitri~V., and Mehrotra, Sharad.
\newblock Context-assisted face clustering framework with human-in-the-loop.
\newblock \emph{International Journal of Multimedia Information Retrieval},
  3\penalty0 (2):\penalty0 69--88, 2014.
\newblock ISSN 2192-6611.
\newblock \doi{10.1007/s13735-014-0052-1}.
\newblock URL \url{http://dx.doi.org/10.1007/s13735-014-0052-1}.

\end{thebibliography}
